\documentclass[runningheads]{llncs}

\usepackage{eccv}

\usepackage{threeparttable}
\usepackage{amsmath}
\usepackage{mathtools}
\usepackage{wrapfig,lipsum}
\usepackage{comment}

\usepackage[ruled,vlined]{algorithm2e}
\definecolor{commentcolor}{RGB}{110,154,155}   % define comment color
\newcommand{\PyComment}[1]{\ttfamily\textcolor{commentcolor}{\# #1}}  % add a "#" before the input text "#1"
\newcommand{\PyCode}[1]{\ttfamily\textcolor{black}{#1}} % \ttfamily is the code font

\usepackage{eccvabbrv}

\usepackage{graphicx}
\usepackage{booktabs}

\usepackage[accsupp]{axessibility}  % Improves PDF readability for those with disabilities.

\usepackage[pagebackref,breaklinks,colorlinks,citecolor=eccvblue]{hyperref}
\usepackage{orcidlink}

\makeatletter
\newcommand{\figcaption}[1]{\def\@captype{figure}\caption{#1}}
\newcommand{\tblcaption}[1]{\def\@captype{table}\caption{#1}}
\makeatother

\begin{document}

\title{Mixed-precision Supernet Training from Vision Foundation Models using Low Rank Adapter} 

\titlerunning{Abbreviated paper title}

%\author{First Author\inst{1}\orcidlink{0000-1111-2222-3333} \and
%Second Author\inst{2,3}\orcidlink{1111-2222-3333-4444} \and
%Third Author\inst{3}\orcidlink{2222--3333-4444-5555}}
\author{Yuiko Sakuma, Masakazu Yoshimura, Junji Otsuka, \\Atsushi Irie, Takeshi Ohashi}

%\authorrunning{F.~Author et al.}
\authorrunning{Y.~Sakuma et al.}

%\institute{Princeton University, Princeton NJ 08544, USA \and
%Springer Heidelberg, Tiergartenstr.~17, 69121 Heidelberg, Germany
\institute{Sony Group Corporation, Tokyo, Japan \\
\email{\{Yuiko.Sakuma, Masakazu Yoshimura, Junji.Otsuka,\\ Atsushi.Irie, Takeshi.A.Ohashi\}@sony.com}
}

\maketitle

\begin{abstract}
  Compression of large and performant vision foundation models (VFMs) into arbitrary bit-wise operations (BitOPs) allows their deployment on various hardware. We propose to fine-tune a VFM to a mixed-precision quantized supernet. The supernet-based neural architecture search (NAS) can be adopted for this purpose, which trains a supernet, and then subnets within arbitrary hardware budgets can be extracted. However, existing methods face difficulties in optimizing the mixed-precision search space and incurring large memory costs during training. To tackle these challenges, first, we study the effective search space design for fine-tuning a VFM by comparing different operators (such as resolution, feature size, width, depth, and bit-widths) in terms of performance and BitOPs reduction. Second, we propose memory-efficient supernet training using a low-rank adapter (LoRA) and a progressive training strategy. The proposed method is evaluated for the recently proposed VFM, Segment Anything Model, fine-tuned on segmentation tasks. The searched model yields about a 95\% reduction in BitOPs without incurring performance degradation.
\end{abstract}

\section{Introduction}
\label{sec:intro}

Segmentation is one of the fundamental vision tasks, which has been widely applied in real-life services such as surveillance and autonomous driving. 
In the era of artificial general intelligence (AGI), large, performant vision foundation models (VFMs) are available. 
For example, along with other powerful models (\cite{zou2024segment,wang2024hierarchical}), the Segment Anything Model (SAM) \cite{kirillov2023segment} is one of the first foundation models for image segmentation. 
SAM is pre-trained on a broad SA-1B dataset that enables powerful generalization and can be used for zero-shot, as well as fine-tuned for a range of downstream segmentation tasks. 
However, SAM's parameter size and the number of bit-wise operations (BitOPs) are enormous (e.g., $\sim3000\mathrm{T}$ BitOPs for the image encoder).
Like SAM, because the BitOPs of VFMs are typically large, BitOPs reduction is required for deploying them to hardware.
Furthermore, compressing VFMs to arbitrary BitOPs enables their deployment to various devices.

Quantization and neural architecture search (NAS) are promising approaches for compressing VFMs to arbitrary BitOPs. 
Recently, mixed-precision quantization methods \cite{dong2019hawq, wang2019haq, wu2018mixed} have been explored, optimizing the quantization bit-widths for each layer or operation.
Supernet-based NAS methods \cite{cai2019once, yu2020bignas,chu2021fairnas, guo2020single, chen2021autoformer, chen2021searching} decouple model training and search, resulting in notable efficiencies in search and final model performance. 
The supernet contains various subnets and is trained in a weight-sharing manner.
In particular, their efficiency stems from training the supernet only once, allowing scalability to different computational budgets without requiring subsequent retraining. 

In this paper, we focus on combining two complex search spaces of mixed-precision quantization and supernet-based NAS to reduce the BitOPs of VFMs.
We fine-tune a pre-trained VFM into a mixed-precision quantized supernet.
However, we encountered two challenges while fine-tuning a VFM into a mixed-precision quantized supernet.
First, we observed that \emph{certain operators commonly chosen in ViT-based search spaces (i.e., embed dimensions) perform poorly when fine-tuning a VFM into a supernet}.
Our experimental observations suggest that search spaces need to be specially designed for VFMs.
However, previous papers have not discussed the effective supernet search space design for VFMs.
Second, a \emph{large memory cost} is required for supernet training.
The mixed-precision quantized supernet \cite{bai2021batchquant} requires training a large search space with the weight-sharing supernet and the quantization parameters of LSQ+ (i.e., step size and offset). 
Moreover, VFMs have a large number of float-point operations (FLOPs) and require large runtime memory for training.
For example, SAM's FLOPs are approximately 2900G for the ViT-L image encoder and require at least 48G memory GPUs for training the mixed-precision search space.
In many cases, memory-intensive GPUs are not available, and having a small runtime memory is preferred. 

Therefore, we propose an efficient and effective method to fine-tune a VFM into a mixed-precision quantized supernet. 
First, we study the \emph{effective search space design} for fine-tuning a VFM.
We compare the different operators (i.e., input resolution, feature size, width, depth, and bit-widths) for SAM-based search space and discuss the trade-offs between performance and BitOPs. 
Second, we propose a \emph{memory-efficient supernet training} method using low-rank adapters (LoRA) \cite{hu2021lora}.
Instead of training the whole parameters, we only train the LoRA decomposed weights and LSQ+ parameters.
However, simply adapting LoRA degrades the performance.
We hypothesize that this is due to the gradient conflict \cite{gong2021nasvit} (supernet weight updates differ within different subnets) and lack of representation capacity.
To address this, we introduce the multi-path architectures using LoRA, named selective and multiplex methods. 
The selective method switches the paths according to the layer bit-widths to improve the representation ability of each bit-width.
The multiplex method trains the base LoRA module and the additional modules bounded with the layer's bit-width for further improving the supernet capacity.
Furthermore, we propose a progressive training strategy to improve the ultra-low bit-width (i.e., 2, 3, and 4-bits) subnets.
The proposed method is evaluated for semantic and instance segmentation tasks, where the proposed method yields about 1.69\% and 3.12\% improvement over QFA with 18\% memory reduction for ADE20k \cite{zhou2017scene} and COCO \cite{lin2014microsoft} datasets, respectively.
The searched subnet yields about a 95\% reduction in BitOPs without incurring performance degradation.

\section{Related Works}

\subsection{Mixed-Precision Quantization}
Although quantization is effective for reducing BitOPs, and ultra-low bit-width quantization methods have been studied, it is known that quantizing certain layers can degrade performance.
To address this issue, many quantization methods introduce mixed-precision by manually selecting specific layers (such as the first layer or batchnorm layers) to be FP32 or high bit-widths like 8-bits.
To automate the decision, mixed-precision methods \cite{van2020bayesian, wang2019haq, chen2021towards} have been proposed.
In this paper, we achieve mixed-precision quantization by integrating bit-widths in the search space and jointly searching for the bit-widths with the network architecture.

\subsection{Supernet-based NAS}

\subsubsection{ViT-based supernet.}
Early works of supernet-based NAS such as OFA \cite{cai2019once} and BigNAS \cite{yu2020bignas} have explored the convolutional neural network (CNN)-based search space and achieved automation of designing efficient architectures. 
Recently, with the emergence of Vision Transformers (ViTs), the ViT-based search space has been explored. 
For example, HR-NAS \cite{ding2021hr}, UniNet \cite{liu2022uninet}, and ElasticViT \cite{tang2023elasticvit} propose a hybrid search space of CNNs and Transformers.
AutoFormer \cite{chen2021autoformer} is one of the first works to design the search space of ViTs.
AutoFormer trains a supernet with a search space that includes different embedding dimensions, depth, MLP ratio, and the number of heads.
It achieves to find competitive architectures compared to ViT-based models.
S3 \cite{chen2021searching} extends AutoFormer by further searching the search space.
Similarly, ViTAS \cite{su2022vitas} and ShitNAS \cite{zhang2023shiftnas} follow AutoFormer's search space.
They propose subnet sampling methods to improve supernet training.
In addition to AutoFormer's search space, ViT-ResNAS \cite{liao2021searching} proposes spatial reduction, which reduces the spatial feature size for each ViT stage.
In this paper, we analyze different candidate operators of embedding dimensions, depth, spatial feature size, input resolution, and bit-widths.
Although elastic resolution is a common practice in the CNN-based search space, it has not been explored much for ViTs because it requires dimension reduction of positional encoding.
In addition, no previous works have addressed the mixed-precision quantized search space.

For training a ViT-based supernet, the gradient conflict is a known problem\cite{gong2021nasvit}.
ElasticViT \cite{tang2023elasticvit} addresses the gradient conflict issue by proposing a conflict-aware subnet sampling method, which samples similar FLOPs subnets together to update the supernet weights.
NASViT proposes a switchable scaling layer, inserted after the self-attention and MLP layers in each Transformer layer.
The scaling layer has separate learnable parameters to perform element-wise multiplication on the outputs from different subnets.
Specifically, the scaling layer's weights are selected depending on the embedding dimension size of the subnet.
The architecture of the multiplication layer is similar to adapters \cite{houlsby2019parameter} which perform the linear transformation.
However, NASViT trains all supernet weights unlike adapters, which freeze the pre-trained weights and only fine-tune the adapter layer weights.
Similar to NASViT, we propose additional trainable layers to avoid the gradient conflict issue and improve the supernet performance.
Our method adapts a switchable architecture like NASViT but freezes the pre-trained weights and only fine-tunes the LoRA weights and LSQ+ parameters for memory efficiency.
Furthermore, we propose multi-path architectures using LoRA to boost performance.

\subsubsection{Mixed-precision quantized supernet.}
Because quantization-aware training (QAT) and supernet training are both challenging, previous approaches have focused on limited conditions. 
For instance, while SPOS \cite{guo2020single} addresses mixed-precision QAT and supernet training, re-training is required for the retrieved subnets.
OQA \cite{shen2021once} trains a quantization-aware supernet but adopts a fixed-precision quantization setting and does not consider mixed-precision quantization.
Only QFA \cite{bai2021batchquant} deals with both QAT and supernet training for the mobilenet-based supernet without re-training after training the supernet.
Specifically, the authors combine the supernet with the learnable step size quantization(LSQ+ \cite{bhalgat2020lsq+}) by attaching the paths of all candidate bit-widths.
Furthermore, they propose the BatchQuant technique, which adaptively re-calculates the learnable step sizes using batch statistics to stabilize the training.
No prior works exist that consider the ViT-based search space.
Moreover, QFA requires a large memory cost for optimizing both supernet weights and LSQ+ parameters.

\subsection{Parameter-Efficient Adapters}
For fine-tuning large language models (LLMs), adapters \cite{houlsby2019parameter} have been introduced, which freeze the original weights and only fine-tune the additional layers.
To address the problem of inference latency in adapters, LoRA \cite{hu2021lora} introduces trainable rank decomposition matrices.
LoRA modifies the forward pass as $h=W_0x + \Delta Wx = W_0x + BAx$, where $W_0 \in \mathbb{R}^{d\times k}$ is the frozen pre-trained weights and $B \in \mathbb{R}^{d\times r}$ and $A \in \mathbb{R}^{r\times k}$ represent the decomposed weights. 
After training, the trained weights can be merged as $W = W_0 + \Delta W$.
Here, $r$ is a small integer.
They apply Gaussian initialization for $A$ and zero for $B$, so $\Delta W$ is initialized as zero at the beginning of training, which stabilizes the training.
Although LoRA was originally proposed for LLMs, successor works apply LoRA for vision tasks such as KAdapter \cite{he2023parameter} and Fact \cite{jie2023fact}.
Although we have adopted KAdapter for fine-tuning SAM with segmentation tasks in preliminary experiments, LoRA performed better.
Rather than proposing different rank decomposition designs, we use regular LoRA and focus on evaluating the effectiveness of parameter-efficient fine-tuning on mixed-precision quantized supernet training.

In this paper, we propose a quantization-aware LoRA to train the mixed-precision search space.
Recent studies \cite{dettmers2024qlora, xu2023qa} have introduced quantization to LoRA.
QLoRA \cite{dettmers2024qlora} quantizes the pre-trained weights $W_0$ to NF4 (a highly squeezed type of floating point numbers) and trains $A$ and $B$ with FP32.
However, since the merged weight, $W$, is FP32, QLoRA cannot be directly applied to our mixed-precision training where the merged weights need to be in arbitrary bit-widths.
QA-LoRA \cite{xu2023qa} represents both $W_0$ and $W$ as INT4.
However, QA-LoRA requires performing channel-wise quantization, which is not practical.
Thus, we propose a general form of quantization-aware LoRA which is applicable for layer-wise quantization.
We use fake quantization where $W_0$ and $W$ are FP32 and quantized, respectively.

\section{Proposed Method}

\subsection{Problem Formulation}
We propose an efficient mixed-precision supernet-based NAS method, that fine-tune a VFM and applies it to segmentation tasks.
Let $f(x, a; \theta)$ be a supernet with input $x$ and $a$ be a subnet, weight $\theta$ that is initialized by pre-trained weight, and $y$ be the output.
In this study, our search space is only the backbone.
Specifically, the supernet is fine-tuned for the target task by computing the optimized shared weights $\theta^*$, according to:
\begin{equation}
    \label{eq:supernet fine-tuning}
    \begin{split}
        &\underset{\theta}{\mathrm{min}} 
        \mathbb{E}_{a\sim U(\mathcal{A})}
        \left[\frac{1}{N} \sum^N_i \mathcal{L}(f(x_i, a; \theta), y_i) \right], \,  (x_i, y_i) \in \mathcal{D}_{\mathrm{trn}}
    \end{split}
\end{equation}
where $\mathcal{A}$ and $\mathcal{D}_{\mathrm{trn}}$ are the search space and training dataset, respectively.

After training the supernet, the optimal subnet $a^*$ is searched that yields the best accuracy on the target task for a given FLOPs budget.
In this study, the validation loss was used as the proxy because evaluating test accuracy requires a long computation time on segmentation tasks.
The search process can be formulated as follows:
\begin{equation}
    \label{eq:search}
    \begin{split}
        %&a^* = 
        %\arg \underset{a \in \mathcal{A}}{\mathrm{min}} \, \frac{1}{N_{\mathrm{sub}} \sum^{N_{\mathrm{sub}, i}} \\
        %\mathcal{L}(f(x_i, a; \theta^*), y_i), (x_i, y_i) \in \mathcal{D}_{\mathrm{val}} \\
        &a^* = 
        \arg \underset{a \in \mathcal{A}}{\mathrm{min}}  \, \frac{1}{N} \sum^N_i \mathcal{L}(f(x_i, a; \theta^*), y_i),  \, (x_i, y_i) \in \mathcal{D}_{\mathrm{val}} \\
        &s.t. \: \mathrm{BitOPs}(a) \leq \tau \\
     \end{split}
\end{equation}
where $\mathrm{BitOPs}(a)$, $\tau$, $\mathcal{D}_{\mathrm{trn}}$ are the BitOPs of subnet $a$, BitOPs budget, and the validation dataset, respectively.
Although BitOPs are used in the experiment, other resource constraints such as latency and memory size can also be used.
We adopt the evolutionary search \cite{real2019regularized} to search for the best-performing architectures within the given resource constraint.
Note that the subnet is evaluated without re-training after the search.
The details of the search space design and the proposed LoRA-based method are provided in Section \ref{sec:proposed method, search space design} and \ref{sec:proposed method, lora}, respectively.

\subsection{Search Space Design} 
\label{sec:proposed method, search space design}

\begin{figure}[tb]
\begin{tabular}{cc}
  \begin{minipage}{.55\textwidth}
      \captionsetup{justification=centering}
      \tblcaption{Candidate operators}
      \label{tab:search space}
      \centering
      \begin{threeparttable}
      \begin{tabular}{@{}lc@{}}
        \toprule
        Candidate operators & Search space\\
        \midrule
        {\bf Resolution}  & $[512, 1024; 64]$\tnote{*}\\
        Spatial feature & $[32, 64; 4]$\tnote{*}\\
        {\bf Depth (stage-wise)} & $[4, 5, 6]$\\
        %Depth (network-wise) & $[20, 24; 1]$\tnote{*}\\
        Embed dimension & $[896, 960, 1024]$\\
        {\bf MLP ratio} & $[3.0, 3.5, 4.0]$\\
        \begin{tabular}{l}{\bf Bit-width} \\(activation, weight) \end{tabular}
        & $[2, 3, 4, 8, 32]$\\
      \bottomrule
      \multicolumn{2}{l}{\small Bold font indicates the selected operators.} \\
      \end{tabular}
      \begin{tablenotes}\footnotesize
        \item[*] $[\mathrm{min, max; step size}]$
      \end{tablenotes}
      \end{threeparttable}
  \end{minipage}

  \begin{minipage}{.45\textwidth}
      \centering
      \includegraphics[height=4.0cm]{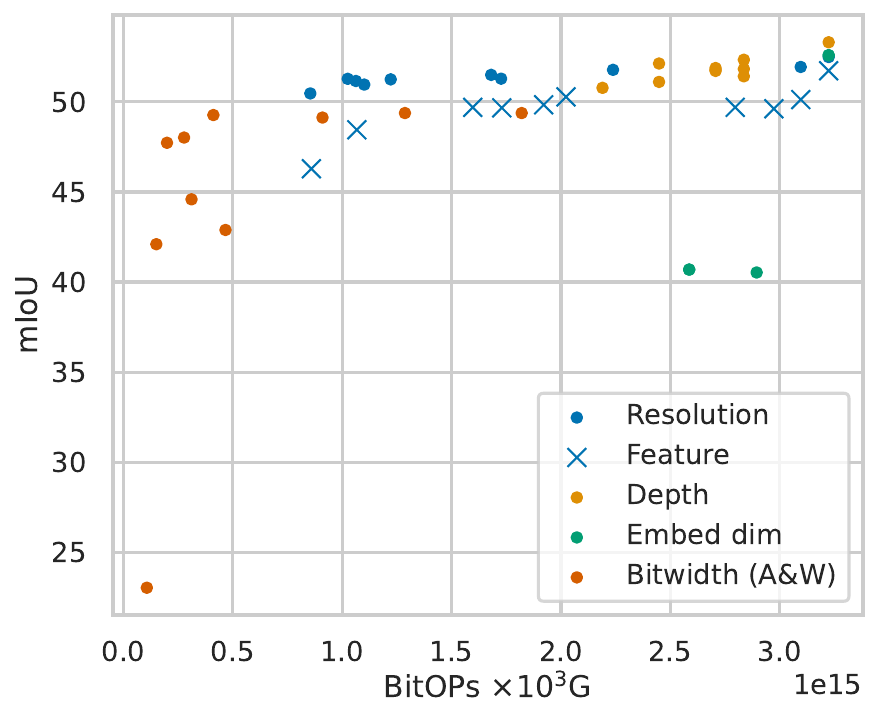}
      \caption{Performance of subnets extracted from the trained supernet of different search spaces}
      \label{fig:operators}
  \end{minipage}
\end{tabular}
\end{figure}

%For training a mixed-precision quantize supernet from SAM, 
We analyze a wide range of operator candidates, including depth, embed dimensions, MLP ratio (adapted in Autoformer \cite{chen2021autoformer}), spatial feature (adapted in ViT-ResNAS \cite{liao2021searching}), input resolutions, and bit-widths.
Input resolutions are not usually selected as the search space in ViT-based NAS although it is common for CNN-based NAS.
This is because the size of the positional encoding is fixed to the largest.
We propose to use linear interpolation of positional encoding to enable elastic resolution for ViTs.
%(ablation studies are provided in the supplementary materials).
For consistency with the input resolution, only the spatial feature after patch embedding is adjusted.
For elastic depth, although the last few layers in the networks are dropped in prior works \cite{chen2021autoformer}, the last $n$ layers in each stage are dropped because segmentation models often use the stage-wise features as the input of heads.
The number of heads is omitted from the candidate operators because the elastic number of heads does not work with SAM implementation.
The candidate operators are elaborated in \cref{tab:search space}.

In the experiment, the supernet was fine-tuned from SAM for each candidate operator's search space for 160k iterations on the ADE20k dataset and semantic segmentation task.
The same experimental setting for QFA* was used (see \ref{sec:experiment, settings} for details).
During training, only the target operators were elastic, and the other operators were set to the largest option.
For elastic bit-widths, the resolution size was set to 768 due to resource limitations.
The bit-widths of the Mask2Former heads were fixed to FP32.
After training, the subnets were randomly sampled from the trained supernet and evaluated as shown in \cref{fig:operators}.
The analysis of the MLP ratio is not provided below because the change in performance was not significant.

\subsubsection{Resolution and spatial feature.} From \cref{fig:operators}, elastic resolution performs better than the spatial feature by about 1.87\% on average.
The performance drop is larger for the elastic feature, especially for smaller subnets; reducing the feature size may directly induce a gap compared to the larger features learned in the pre-trained models and degrade the performance.
The change in resolution induces less difference in the patch embedding and performs better.

\subsubsection{Depth and embed dimension.} The performance drop of elastic depth is relatively smaller than that of the embed dimension.
For instance, the smallest subnet yields an mIoU of 50.78 while reducing BitOPs by about 32\%.
On the contrary, the performance drop for the elastic embed dimension is large; the mIoU is smaller by about 10\% compared to the elastic depth for similar BitOPs.
One of the reasons is that the embed dimension changes the width of the network globally because of the skip connection.
We have also changed the width stage-wise by using a linear transform of the width between stages, but still observed a large performance drop.
We hypothesize that the channel-wise representations are important in VFMs.
Interestingly, this aligns with the findings from recent studies on LLM and large vision models \cite{radford2019language, el2024scalable}, which suggest that a large width is more important than a large depth.

\subsubsection{Bit-width.} Some mixed-precision subnets perform well with significant reductions in BitOPs such that the top six subnets in \cref{fig:operators} yield mIoU between 47.73 and 49.38, with a reduction of 89\% in BitOPs.
Although some subnets perform worse (e.g., the 2-bit subnet with mIoU of 23.03), by enabling the search for the best-performing mixed-precision subnets, it is possible to find good-performing subnets within specific hardware budgets.

\subsubsection{Summary.} The elastic resolution, depth (stage-wise), MLP ratio, and bit-width yield better trade-offs between performance and BitOPs compared to elastic feature and embed dimension.
For fine-tuning pre-trained SAM, operators that affect features less may achieve small performance degradation.
We focus on the well-performing operators indicated in bold font in \cref{tab:search space}, which covers a large search space (about $10^8$ candidate subnets in total) in terms of BitOPs.

\subsection{Supernet Fine-Tuning with Low-Rank Adapter} 
\label{sec:proposed method, lora}

We propose a memory-efficient and effective fine-tuning method using LoRA to fine-tune a VFM into a mixed-precision quantized supernet.
We use the recently proposed LoRA to fine-tune a supernet from a VFM where fine-tuning is more complex than adaptations for a single network as in the usual case.
Surprisingly, we observed that simply fine-tuning SAM with regular LoRA (\cref{fig:proposed LoRA-a}) yields similar performance to full parameter training for some subnets as detailed in the experiment section.
However, regular LoRA struggles with training ultra-low bit-width subnets.
We hypothesize that this is caused by the gradient conflict \cite{gong2021nasvit} between subnets with larger and smaller bit-widths and the lack of representation capacity.
To address this issue, (1) multi-path LoRA-based architectures are proposed.
Since the proposed LoRA needs to train a quantized search space, (2) a general form of quantization-aware LoRA using fake quantization is proposed.
Finally, to boost the performance of ultra-low bit-width subnets, (3) the progressive training method is proposed.

\subsubsection{Multi-path LoRA-based Architectures.}
\label{sec:proposed method, multi-path lora}

\begin{figure}[tb]
  \centering
  \begin{subfigure}{0.20\linewidth}
    \includegraphics[width=\linewidth]{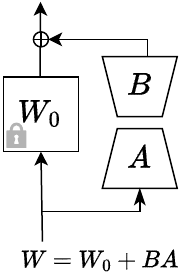}
    \caption{Regular LoRA}
    \label{fig:proposed LoRA-a}
  \end{subfigure}
  \hfill
  \begin{subfigure}{0.35\linewidth}
    \includegraphics[width=\linewidth]{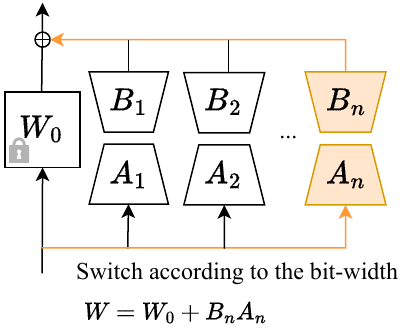}
    \caption{Selective method}
    \label{fig:proposed LoRA-b}
  \end{subfigure}
  \hfill
  \begin{subfigure}{0.40\linewidth}
    \includegraphics[height=3.5cm]{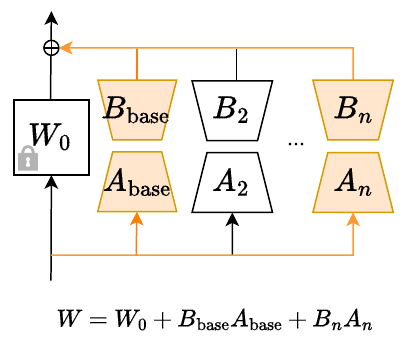}
    \caption{Multiplex method}
    \label{fig:proposed LoRA-c}
  \end{subfigure}
  \caption{The proposed LoRA-based architecture.
  (a) The regular LoRA freezes the pre-trained weight $W_0$ and only fine-tunes the low-rank decomposition weights of $A$ and $B$.
  (b) The selective method switches the decomposed weights $A_n$ and $B_n$ according to the layer bit-widths.
  (c) The multiplex method always trains the base weights $A_{\mathrm{base}}$ and $B_{\mathrm{base}}$.
      For low bit-width layers. Further, the additional weights bounded with the layer's bit-width $A_n$ and $B_n$ are trained.}
  \label{fig:proposed LoRA}
\end{figure}

We propose three LoRA-based architectures presented in \cref{fig:proposed LoRA}.
First, \cref{fig:proposed LoRA-a} presents the regular LoRA where $W_0$ is frozen and only the LoRA decomposed weights $A$ and $B$ are fine-tuned.
Second, the selective method (\cref{fig:proposed LoRA-b}) switches the LoRA modules according to the layer's activation or weight bit-widths, and only the selected weights $A_n$ and $B_n$ are updated.
LoRA weights are initialized separately.
This is similar to NASViT \cite{gong2021nasvit} where the adapter-like layer is switched according to the embed dimension.
The difference to NASViT is using LoRA, and only LoRA decomposed weights are fine-tuned.
Finally, the multiplex method (\cref{fig:proposed LoRA-c}) always trains the base LoRA weights $A_{\mathrm{base}}$ and $B_{\mathrm{base}}$.
For lower bit-width subnets, the additional weights bounded with the layer's bit-widths ($A_n$ and $B_n$) are trained.
For larger bit-widths, only the base LoRA weights are trained.
In the supplementary materials, the combinations of the number of LoRA modules and bit-width assignments are detailed as well as the pseudo-code.
The multiplex design improves the capacity of the network for ultra-low bit-width subnets where the subnets' outputs change more and additional fine-tuning may be required.
Furthermore, this design is related to the bit inheritance methods such as OQA \cite{shen2021once}.
OQA initializes the low bit-width supernets by inheriting the weights of larger bit-widths.
This inheritance technique improves the performance of subnets with smaller bit-widths because the optimal weights are similar between adjacent bit-widths.
The multiplex architecture explicitly models the difference to the subnets of higher bit-widths and improves the accuracy for subnets with lower bit-widths.
Note that LoRA is adapted only during training, and for inference, it is merged with the pre-trained weight $W_0$ and LoRA weights are removed.

\subsubsection{The General Form of Quantization-Aware LoRA.}
\label{sec:proposed method, qa-lora}

Before formulating the proposed quantization-aware LoRA, the common notation for quantization is introduced.
In this study, the LSQ+ technique \cite{bhalgat2020lsq+} is used, which applies uniform asymmetric quantization.
Let $x$ be an FP32 input with a range of $(x_{\mathrm{min}}, x_{\mathrm{max}})$.
For $b$ bit-width precision, the integer quantized value $x_{\mathrm{q}}$ has a range of $[n, p] = [0, 2^b - 1]$.
Then, two trainable parameters, scale $\mathrm{\Delta}$ and zero-point $z$, are derived to map the FP32 values to integers.
The scale is the step size of the quantizer and the zero-point maps the original zero value.
The quantization procedure can be formulated as follows:
\begin{align}
  \mathrm{\Delta} & = \frac{x_{\mathrm{max}} - x_{\mathrm{min}}}{p - n}, \: z=\mathrm{clamp} \left( \left\lfloor \frac{x_{\mathrm{min}}}{\mathrm{\Delta}} \right\rceil n, p \right) \\
  x_{\mathrm{q}} &= \left \lfloor \mathrm{clamp} \left( \frac{x}{\mathrm{\Delta}} + z, n, p \right) \right \rceil \\
  \hat{x} &= \left( x_{\mathrm{q}} - z \right) \mathrm{\Delta} 
  \label{eq:lsq+}
\end{align}
where $\lfloor \cdot \rceil$ denotes the round function and $\mathrm{clamp}(\cdot)$ denotes the clamp operation to fall between $n$ and $p$.
The scale $\mathrm{\Delta}$ and zero-point $z$ are trained jointly with the supernet weights.
According to LSQ+, for weight quantization, only $\mathrm{\Delta}$ is trained.
For activation quantization, both $\mathrm{\Delta}$ and $z$ are trained.

\begin{algorithm}[tb]
\caption{General form of quantization-aware LoRA in PyTorch-like style}\label{alg:qa-lora}
\SetAlgoLined
    \PyComment{W\_0, A, B: pre-trained weights, LoRA decomposed weights A and B} \\
    \PyComment{x: input} \\
    \PyCode{def qalora\_forward(x, W\_0, A, B):} \\
    \Indp
        \PyCode{W\_tilde = pre\_quantization(W\_0, A, B)} \\
        \PyCode{y = x @ W\_tilde} \\
        \PyCode{y += x @ A @ B * scaling} \\
        \PyCode{return y} \\
    \Indm
    \PyCode{} \\
    \PyCode{def pre\_quantization(W\_0, A, B):} \\
    \Indp
        \PyCode{BA = A @ B} \\
        \PyCode{W\_tilde = quantize(W\_0 + BA) - BA} \\
        \PyCode{return W\_tilde} \\
    \Indm
\end{algorithm}

As described in the previous section, we use LoRA for mixed-precision quantized supernet training.
Since the LoRA weights are merged into $W_0$ during inference, the merged weights need to be quantized.
Although quantization-aware LoRA, QA-LoRA \cite{xu2023qa}, is proposed, this method assumes a channel-wise scale.
In practice, this requires extra memory for hardware implementation during inference, and layer-wise quantization is preferred.
Thus, we propose a general form of quantization-aware LoRA for weight quantization.
Here, the proposed quantization-aware LoRA is explained for the regular LoRA (\cref{fig:proposed LoRA-a}), but the proposed formulation can be directly extended to selective and multiplex methods.
For LoRA, the merged weight $W$ is the sum of the pre-trained weight $W_0$ and decomposed weights $A$ and $B$, where $W = W_0 + BA$.
To guarantee that $W$ is quantized, we introduce $\tilde{W}$, which represents the difference between the quantized weight $\hat{W}$ and $BA$, and is formulated as:
\begin{align}
  \hat{W} &= \left( W_{\mathrm{q}} - z \right) \mathrm{\Delta}, \: W_{\mathrm{q}} = \left \lfloor \mathrm{clamp} \left( \frac{W}{\mathrm{\Delta}} + z, n, p \right) \right \rceil \\
  \Tilde{W} &= \hat{W} - BA.
  \label{eq:lsq+}
\end{align}
For the forward calculation, $\tilde{W}$ is used instead of $W_0$, and we have $y = (\tilde{W} + AB)x$.
During training, fake quantization is used, where all weights $\tilde{W}$, $\hat{W}$, $W_0$, $A$, and $B$, as well as gradients, are processed in FP32.
The proposed general form of quantization-aware LoRA is summarized in \cref{alg:qa-lora}.
Different from QA-LoRA \cite{xu2023qa}, our method is general and can be used for both channel-wise and layer-wise quantization.

\subsubsection{Progressive Training.}
\label{sec:proposed method, progressive training}

In the preliminary experiments, we noticed that training ultra-low bit-width subnets proved to be challenging when using LoRA for fine-tuning throughout all iterations.
This is because LoRA, which allows for small updates from the pre-trained weights, restricts the optimization of the supernet for ultra-low bit-width subnets whose activation is significantly different from those with higher bit-widths.
To address this issue, we introduce a progressive training strategy.
After all the weights of the supernet are fine-tuned for some iterations, the LoRA modules are attached.
For LoRA training, the pre-trained weights are frozen and only LoRA weights are trained.
To enable training with limited memory, we trained the supernet with a smaller resolution (i.e., a maximum of 768) for the full parameter training.
After the full parameter training, the maximum resolution is adjusted to 1024 and fine-tuned with LoRA.
Through experimentation, we have confirmed that this setting allows the subnets with the highest resolution to regain their accuracy.
The idea of progressive training is similar to the progressive shrinking technique \cite{cai2019once}, where the search space is progressively trained from larger to smaller subnets.
We observed that for fine-tuning SAM, the progressive shrinking suffers from training smaller subnets and our proposed strategy is effective for training them.

\section{Experiments}
\subsection{Experimental Settings}
\label{sec:experiment, settings}

\begin{figure}[tb]
  \centering
  \begin{subfigure}{0.48\linewidth}
    %\fbox{\rule{0pt}{0.5in} \rule{.9\linewidth}{0pt}}
    \includegraphics[width=\linewidth]{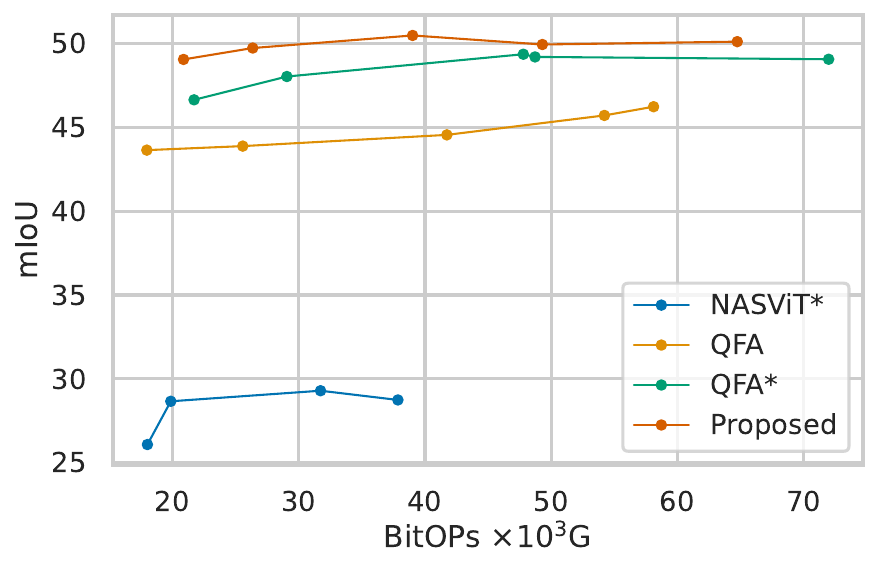}
    \caption{Semantic segmentation/ADE20k}
    \label{fig:mixed-precision-a}
  \end{subfigure}
  \hfill
  \begin{subfigure}{0.48\linewidth}
    %\fbox{\rule{0pt}{0.5in} \rule{.9\linewidth}{0pt}}
    \includegraphics[width=\linewidth]{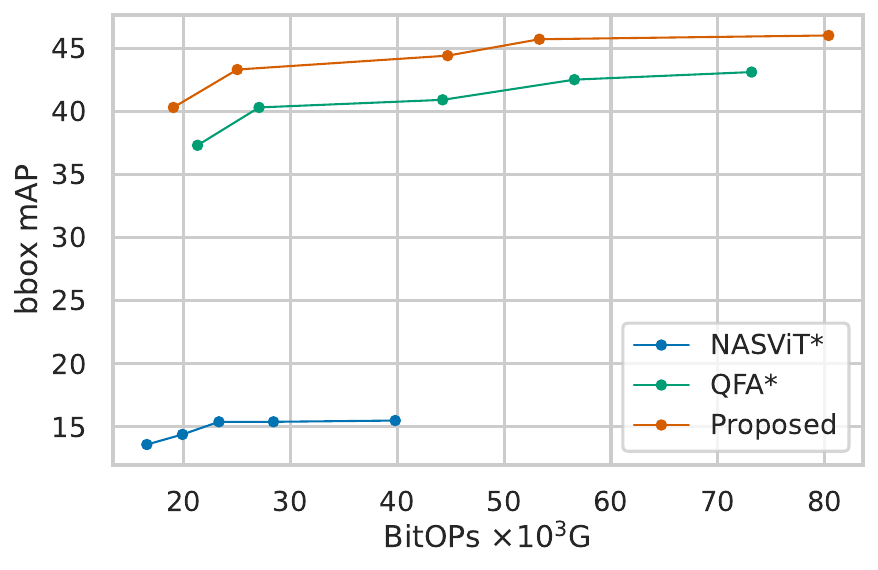}
    \caption{Instance segmentation/COCO}
    \label{fig:mixed-precision-b}
  \end{subfigure}
  \caption{Comparison with state-of-the-art mixed-precision supernet methods for subnets with different BitOPs budget}
  \label{fig:mixed-precision}
\end{figure}

%We fine-tune the pre-trained SAM into the mixed-precision supernet for segmentation tasks.
%The supernet search space is detailed in \cref{tab:search space}.
Only the backbone architecture is searched, and the head architecture is not searched.
The proposed LoRA modules are attached to the query, key, and value projection matrices in the self-attention module and MLP feedforward layers.
The SAM-L image encoder was used as the backbone and searched (see supplementary materials for model details).
The prompt encoder and mask decoder were removed, and the Mask2Former \cite{cheng2022masked} head was attached as the task-specific head.
The target quantization layers are the same as GPUSQ-ViT \cite{yu2023boost}, such that the quantized layers include patch embedding, final linear projection, as well as the feed-forward and linear projection inside each transformer block.
The layernorm layers were not quantized.
Min-max quantization of 8-bits was applied to the head.

The supernet was trained on eight V100 GPUs with 32GB memory and A100 GPUs with 48GB memory for the proposed and comparative methods (QFA and NASViT), respectively.
The same training hyper-parameter settings as Mask2Former \cite{cheng2022masked} were used except for the learning iterations.
For the semantic segmentation task with ADE20k and instance segmentation with COCO, the supernet was fine-tuned for 240k iterations and 80 epochs in total, respectively.
For the proposed progressive training, full parameter training was applied for 120k iterations and 50 epochs, respectively. 
After training, the supernet is searched for the best-performing subnet under the given resource constraint.
The same evolutionary search hyper-parameters as AutoFormer \cite{chen2021autoformer} were used except for the number of epochs (i.e., five instead of 20) due to the resource limits.
Otherwise specified, the multiplex LoRA architecture with five paths is denoted as the proposed method.

\subsection{Comparison with Mixed-precision Quantized NAS}

The proposed method is compared with the state-of-the-art mixed-precision supernet methods, NASViT \cite{gong2021nasvit} and QFA \cite{bai2021batchquant}.
Both NASViT and QFA were trained using the same search space and hyper-parameter settings as the proposed method.
For NASViT, to evaluate the efficacy of the proposed LoRA architecture, only the switchable scaling layer was adapted, and the gradient projection and data augmentation techniques were abbreviated.
Furthermore, because their proposed scaling layer did not converge for our search space, the adapter \cite{houlsby2019parameter} architecture with the same parameter size was used instead and denoted as NASViT*.
The scaling layer was switched according to the layer depth.
Since NASViT* requires a large runtime memory beyond our hardware budget, the resolution search space was limited to a maximum of 768.
QFA was trained with/without the BatchQuant technique and denoted as QFA and QFA*, respectively.
Note that QFA* is the full parameter training version of the proposed method.

\begin{wraptable}{r}{6.5cm}
      \caption{Memory costs during training}
      \label{tab:memory costs}
      \centering
      \begin{tabular}{@{}lc@{}}
        \toprule
        Method & \begin{tabular}{l} Memory\\cost (GB) \end{tabular}\\
        \midrule
        NASViT  &  - \\
        QFA  &  28.19 \\
        Proposed (regular LoRA) & 20.71\\
        Proposed (selective LoRA) & 20.67\\
        Proposed (multiplex LoRA, 3*) & 22.47\\
        Proposed (multiplex LoRA, 4*) & 22.74\\
        Proposed (multiplex LoRA, 5*) & 23.03\\
      \bottomrule
      \multicolumn{2}{l}{\small *The number of LoRA modules} \\
      \end{tabular}
\end{wraptable}

In \cref{fig:mixed-precision-a}, the proposed method is evaluated on the semantic segmentation task using the ADE20k dataset by searching for the best-performing subnets under different BitOPs budgets.
Our proposed method not only reduces memory costs by approximately 18\% (\cref{tab:memory costs}), but also achieves comparable or better performance than the full parameter training, QFA*.
On average, an improvement of about 1.40\% is observed.
The BatchQuant technique of QFA does not perform well compared to the proposed method.
In fact, the proposed method is about 5.06\% better than QFA on average.
This suggests that the BatchQuant technique is more suitable for the CNN-based search spaces.
In ViT-based search spaces, greater variations in activations between different subnets occur compared to the CNN-based ones\cite{gong2021nasvit}, and the batch-wise scale correction degrades the performance.
Furthermore, NASViT* performs poorly when compared to both the proposed method and QFA (the proposed method outperforms by about 21.90\%).
This highlights the effectiveness of the LoRA-based fine-tuning method.
For fine-tuning the pre-trained VFMs, the degree of weight updates is smaller than training from scratch.
Thus, the LoRA-based approach is effective because it avoids making excessively large updates from the pre-trained representations, while still successfully adapting to downstream tasks.
In \cref{fig:mixed-precision-b}, the proposed method is evaluated for the instance segmentation task using the COCO dataset.
Similar to the semantic segmentation task, the proposed method consistently outperforms QFA* and NASViT* by about 3.12\% and 29.08\% on average, respectively.

\subsection{Comparison with Fixed-precision Quantized NAS}

\begin{table}[tb]
      \caption{Comparison between fixed precision results}
      \label{tab:fixed-precision}
      \centering
      \begin{tabular}{@{}lccc|lccc@{}}
        \toprule
        Method & Bit-width & \begin{tabular}{l}BitOPs\\($\times 10^3$G)\end{tabular} & mIoU & Method & Bit-width & \begin{tabular}{l} BitOPs\\($\times 10^3$G)\end{tabular} & mIoU\\
        \midrule
        PSAQ-ViT V2 & W32/A32 & 1063.00 & 49.30 & PSAQ-ViT V2 & W4/A8 & 33.20 & 44.60\\
        GPUSQ-ViT &           & 1188.00 & 48.13 & GPUSQ-ViT & W4/A4 & 19.60 & \textbf{47.86}\\
        \checkmark QFA* &     &  758.16 & 48.21 & \checkmark QFA* &       & 19.91 & 46.17\\
        \checkmark Proposed&  &  747.16 & \textbf{50.25} & \checkmark Proposed &  & 19.37 & 46.59\\
        PSAQ-ViT V2 & W8/A8   &   66.40 & 47.20 & \checkmark QFA*   & W3/A3 & 12.43 & 43.28\\
        GPUSQ-ViT &           &   39.20 & 48.18 & \checkmark Proposed &     & 13.00 & \textbf{46.16}\\
        \checkmark QFA* &     &   50.32 & 48.23 & \checkmark QFA*   & W2/A2 & 6.00 & 28.98\\
        \checkmark Proposed & &   51.82 & \textbf{50.02} & \checkmark Proposed &  & 6.02 & \textbf{30.25}\\
      \bottomrule
      \multicolumn{8}{l}{\small \checkmark indicates the supernet-based methods.} \\
      \end{tabular}
\end{table}

The proposed method is compared for the fixed-precision subnets for both the supernet (QFA*) and stand-alone methods (PSAQ-ViT V2 \cite{li2023psaq} and GPUSQ-ViT \cite{yu2023boost}).
The subnets with the closest BitOPs to stand-alone models with the same bit-width settings were searched for QFA* and the proposed method.
From \cref{tab:fixed-precision}, the proposed method outperforms the comparative methods for 2, 3, 8, and 32-bit settings.
Although GPUSQ-ViT outperforms for the 4-bit setting, it only provides results above 4-bit, and the performance for lower bit-width models is not revealed.
On the contrary, our method can train 2 and 3-bit settings as well.
Thus, without re-training, our method achieves competitive results to stand-alone models.
Furthermore, by comparing the FP32 results and the mixed-precision results (\cref{fig:mixed-precision-a}), the searched model yields approximately a 95\% reduction in BitOPs without incurring performance degradation.

\subsection{Ablation Studies of Proposed LoRA}

\begin{figure}[tb]
  \centering
  \begin{subfigure}{0.48\linewidth}
    %\fbox{\rule{0pt}{0.5in} \rule{.9\linewidth}{0pt}}
    \includegraphics[width=\linewidth]{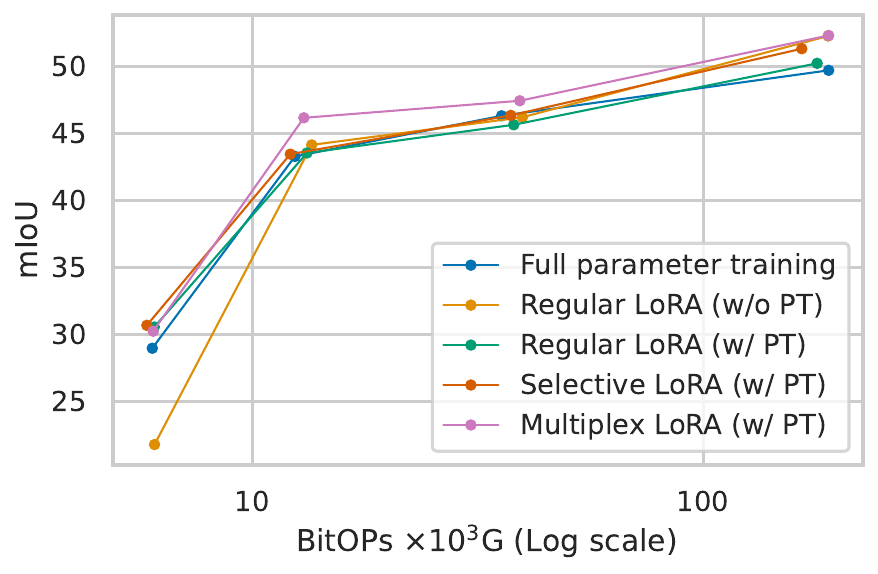}
    \caption{Progressive training (PT) and LoRA designs}
    \label{fig:lora ablations-a}
  \end{subfigure}
  \hfill
  \begin{subfigure}{0.48\linewidth}
    %\fbox{\rule{0pt}{0.5in} \rule{.9\linewidth}{0pt}}
    \includegraphics[width=\linewidth]{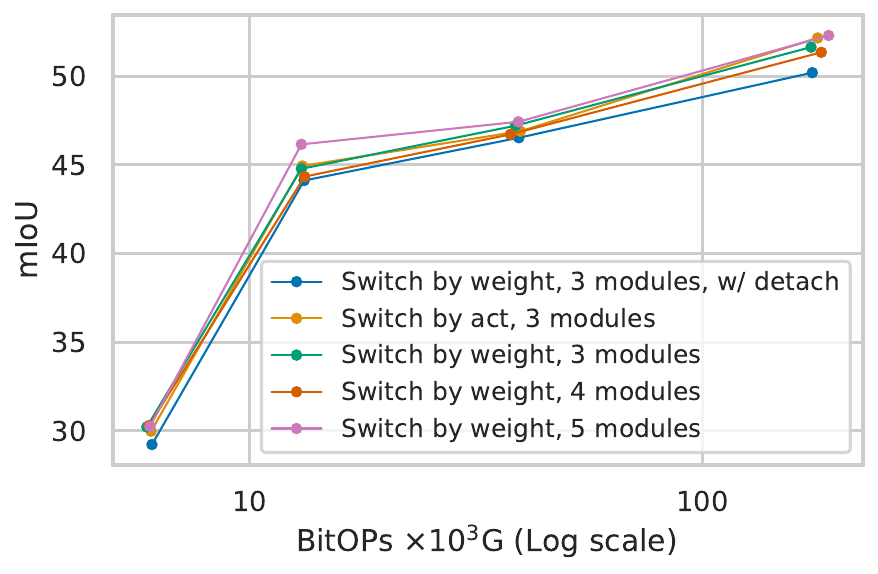}
    \caption{Multiplex LoRA ablations}
    \label{fig:lora ablations-b}
  \end{subfigure}
  \caption{Ablation studies for the proposed LoRA-based architectures}
  \label{fig:lora ablations}
\end{figure}

\subsubsection{Effect of Progressive Training (PT).}
The subnets are searched from the trained supernet for the fixed precisions of 2, 3, 4, and 8-bit for both weight and activation.
As shown in \cref{fig:lora ablations-a}, the proposed PT improves the performance of ultra-low bit-width subnets (i.e., about 8.8\% improvement for the 2/2-bit subnet).
Specifically, we observed that without PT, 2-bit precision subnets perform poorly.
This may be due to the large discrepancy between the outputs of 2-bit and FP32 subnets.

\subsubsection{Comparison of LoRA Designs.}
From \cref{fig:lora ablations-a}, the proposed multiplex LoRA architecture outperforms the full parameter training by about 1.69\% on average.
Furthermore, the selective and multiplex methods outperform the regular LoRA by about 0.42\% and 1.61\% on average, respectively.
The multiplex method performs the best among the three LoRA-based architectures and is 1.19\% better than the selective method on average.
Although the multiplex architecture requires extra memory costs (see \cref{tab:memory costs}), the overhead is small (i.e., about 11.2\% increase to regular LoRA) because the extra layers are also the memory-efficient LoRA architecture.

\subsubsection{Comparison of Multiplex LoRA Designs.}
We compare different settings of the proposed multiplex LoRA architecture shown in \cref{fig:lora ablations-b}.
First, the base LoRA weights are updated only when larger bit-widths are selected (denoted as ``w/ detach'', detailed in the supplementary materials).
This performs worse than without this setting, suggesting that updating all LoRA weights gives extra representation ability of the supernet without harming the performance for the larger bit-width subnets.
Second, the different switching strategy was evaluated; the LoRA modules were switched according to the activation or weight bit-widths of the layers (denoted as ``switch by weight'' or ``act'').
Both performed similarly and we consider that the LoRA modules can be switched by either weight or activation bit-width settings.
Finally, the effect of LoRA module numbers was evaluated.
As the number of LoRA modules increases, the performance improves.
For instance, the mIoU of five module settings is 0.63\% better than three.

\section{Conclusion}
In this paper, we propose an efficient and effective mixed-precision quantized supernet training for pre-trained VFMs using a low-rank adapter, which enables a joint search for architecture and mixed-precision quantization policy.
We provide an analysis of the effective search space design for fine-tuning SAM as a mixed-precision supernet.
For resource-efficient training, we introduce LoRA-based supernet architectures where the multiplex design yields the best performance while allowing training within the 32GB memory limit.
We demonstrate the robust performance of the proposed method for both the mixed- and fixed-precision (including ultra-low bit-widths, 2/3/4-bits) subnets extracted from the trained supernet where it achieves about a 95\% reduction in BitOPs without incurring performance degradation.
Thus, our method contributes to the deployment of VFMs on real-time hardware.

\textbf{Limitations of this work} include the extension of the proposed LoRA-based method for scratch training.
Although this work focuses on the fine-tuning of pre-trained VFMs, by applying memory-efficient scratch training methods like ReLoRA \cite{lialin2023stack}, our method may be applied for scratch training as well.

%\par\vfill\par
%Now we have reached the maximum length of an ECCV \ECCVyear{} submission (excluding references).
%References should start immediately after the main text, but can continue past p.\ 14 if needed.
%\clearpage  % TODO REVIEW/FINAL: This \clearpage needs to be removed from both review and camera-ready versions.

\bibliographystyle{splncs04}
\bibliography{main}

\clearpage
% ---- Supplementary Material ----

%\begin{comment}
\section{Supplementary Material}
\subsection{SAM Image Encoder Details}
The SAM image encoder follows the ViT architecture with 14$\times$14 windowed attention and four equally-spaced global attention blocks.
It uses the input resolution of 1024$\times$1024 obtained by re-scaling the image and padding the shorter side.
The image embedding is 64$\times$64.
The ViT-L architecture is summarized in \cref{tab:ViT-L architecture}.
The image encoder is first pre-trained using the masked autoencoder (MAE).
Then, the image encoder, prompt encoder, and mask decoder are trained on the large segmentation dataset, SA-1B \cite{zou2024segment}.

\begin{table}[h]
      \caption{ViT-L architecture}
      \label{tab:ViT-L architecture}
      \centering
      \begin{tabular}{@{}cccc@{}}
        \toprule
        Depth \: & Hidden size \: & MLP ratio \: & Number of heads\\
        \midrule
        24  & 1024 & 4.0 & 16\\
      \bottomrule
      \end{tabular}
\end{table}

\subsection{Multiplex LoRA Details}
\subsubsection{Design Details.}
In \cref{tab:multiplex LoRA assignment}, the bit-width assignment for the proposed multiplex LoRA (\cref{fig:proposed LoRA-c}) is presented.
Each bracket presents the assigned bit-widths for each LoRA module.
The largest $n$ bit-widths are grouped, and lower bit-widths are assigned to different modules.
The LoRA module, which FP32 is assigned, is the base.

\begin{table}[h]
      \caption{Bit-width assignments for different number of LoRA modules}
      \label{tab:multiplex LoRA assignment}
      \centering
      \begin{tabular}{@{}lc@{}}
        \toprule
        Number of LoRA modules  \: & Bit-widths assignment\\
        \midrule
        3  & $\{[2], [3], [4, 8, 32]\}$\\
        4  & $\{[2], [3], [4], [8, 32]\}$\\
        5  & $\{[2], [3], [4], [8], [32]\}$\\
      \bottomrule
      \end{tabular}
\end{table}

\subsubsection{Experiment Details.}
We evaluated the proposed multiplex LoRA architecture for the different settings.
Specifically, the effect of freezing the base LoRA weights (denoted as ``w/ detach'') is studied.
Under this setting, when the low bit-width ($n$-bits) is selected, the base LoRA weights are frozen and only the corresponding LoRA weights ($A_n$ and $B_n$) are updated.
The PyTorch-like pseudo-code is provided in \cref{alg:detached LoRA}.
Otherwise denoted as ``w/ detach'', the ``no\_grad'' operation is simply removed.

\begin{algorithm}[h]
\caption{Detached version of the proposed multiplex LoRA in PyTorch-like style}
\label{alg:detached LoRA}
\SetAlgoLined
    \PyComment{W\_0: pre-trained weights} \\
    \PyComment{A\_n, B\_n: LoRA decomposed weights for n-bit}\\
    \PyComment{A\_base, B\_base: LoRA decomposed weights for the base LoRA module} \\
    \PyComment{x: input} \\
    \PyComment{bits: bit-width of the layer} \\
    \PyComment{base\_bits: bit-widths assigned to the base LoRA module} \\
    \PyCode{y = x @ W\_0} \\
    \PyCode{if bits in base\_bits:} \\
    \Indp
        \PyCode{y += x @ A\_base @ B\_base * scaling} \\
    \Indm
    \PyCode{else:} \\
    \Indp
        \PyCode{with torch.no\_grad():} \\
        \Indp
            \PyCode{y += x @ A\_base @ B\_base * scaling} \\
        \Indm
    \Indm
    \PyCode{if bits not in base\_bits:} \\
    \Indp
        \PyCode{y += x @ A\_n @ B\_n * scaling} \\
    \Indm
\end{algorithm}

%\end{comment}

\end{document}